\documentclass[runningheads]{llncs}

\usepackage{eccv}

\usepackage{eccvabbrv}

\usepackage{graphicx}
\usepackage{booktabs}
\usepackage{amsmath}
\usepackage{amssymb}
\usepackage{multirow}
\usepackage{colortbl}
\usepackage[ruled,vlined]{algorithm2e}
\definecolor{mygray}{gray}{.85}
\usepackage{hyperref}
\usepackage{tabularx}
\usepackage{enumitem}

\usepackage[accsupp]{axessibility}  

\usepackage{hyperref}

\usepackage{orcidlink}
\usepackage{fontawesome}
\usepackage{marvosym}

\def\netName{MCNav}
\newcommand\blfootnote[1]{%
  \begingroup
  \renewcommand\thefootnote{}\footnotetext{#1}%
  \addtocounter{footnote}{-1}%
  \endgroup
}

\begin{document}

\title{MCNav: Memory-Aware Dynamic Cognitive Map for Zero-shot Goal-oriented Navigation} 

\titlerunning{MCNav}




\author{Jingyu Li\inst{1,2*} \and
Zhe Liu\inst{3*} \and
Wenxiao Wu\inst{4,2} \and Li Zhang\inst{1,2\dag}}
\authorrunning{J. Li et al.}
\institute{
\footnotesize
\textsuperscript{1}Fudan University \hspace{0.4em}
\textsuperscript{2}Shanghai Innovation Institute\\
\textsuperscript{3}University of Hong Kong \hspace{0.4em}
\textsuperscript{4}Huazhong University of Science and Technology
}

\maketitle

\blfootnote{
\noindent $*$ Equal contribution. \\
$\dag$ Corresponding author.
}

\begin{abstract}
Navigating to instance-level targets in complex environments is a challenging problem. Many existing zero-shot methods achieve strong performance by modeling the entire environment and leveraging large language models for scene understanding. However, such strategies primarily focus on exploring new regions while lacking a deeper exploitation of information from previously explored areas. 
Consequently, when targets are missed or misidentified within previously visited regions, navigation failures occur frequently.
To address these limitations, we propose \netName{}, a memory-aware navigation framework with a dynamic cognitive map. This map stores efficiently queryable information about relevant objects in explored areas. Building on this memory structure, \netName{} introduces two memory-aware exploration strategies: goal re-validation, which re-assesses previously seen objects to correct matching failures, and missed goal re-exploration, which estimates the likelihood that a target is present in an explored region from contextual cues. These strategies are further stabilized by a blacklist mechanism to prevent repeated errors and a  double-check mechanism for high-confidence confirmation.
We evaluate \netName{} on the HM3Dv1 and HM3Dv2 datasets across three different tasks, where it achieves state-of-the-art performance, particularly on the instance-level goal navigation task. 
\end{abstract}

\section{Introduction}

Goal-oriented navigation, a fundamental task in embodied AI~\cite{li2025imagidrive,liu2025drivepi,li2026sgdrive,fast-wam,lingbot-va2026,bi2025motus} and robotics, focuses on enabling an autonomous agent to reach a specified goal within an unknown environment. 
It can be categorized into three primary tasks based on the form of goal specification: Object goal navigation~(ON), which guides the agent using an object category; Instance Image Navigation~(IIN), where the goal is defined via an instance-level reference image; and Text goal navigation~(TN), which specifies the target through text descriptions.
Driven by recent advances in large language models~(LLMs) and vision-language models~(VLMs), various zero-shot navigation approaches~\cite{yin2025unigoal,apexnav,cai2024bridging,zhou2023esc, busch2025one} have emerged to tackle the aforementioned tasks, achieving promising performance. 

\begin{figure}[tb]
    \centering
    \includegraphics[width=\linewidth]{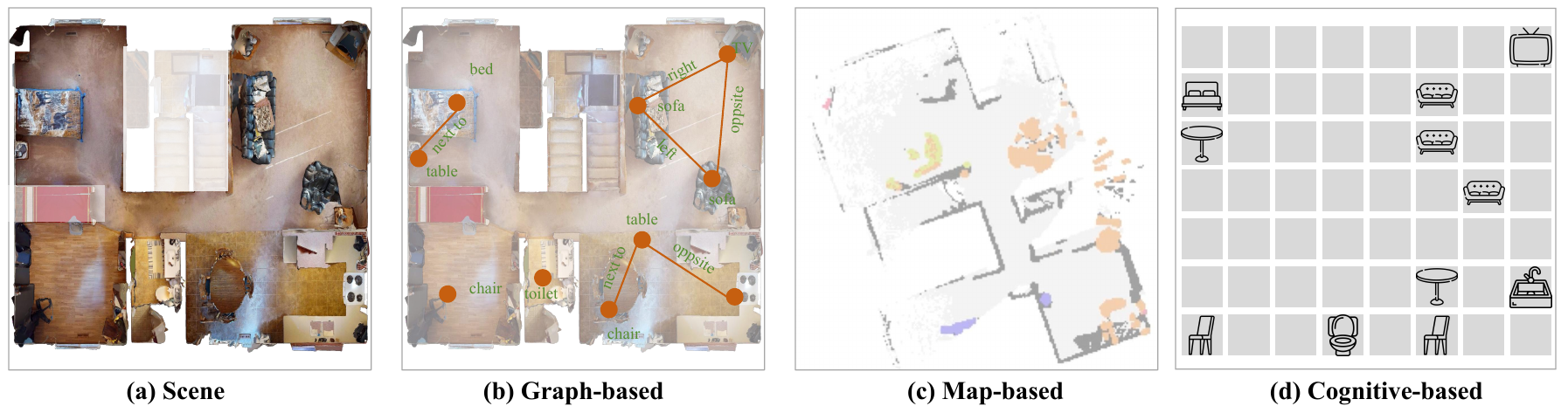}
    \caption{Different scene representation paradigms adopted in zero-shot goal-oriented navigation: (a) the raw scene; (b) graph-based methods that construct node–edge structures; (c) map-based methods that build semantic occupancy maps; and (d) our cognitive map representation, which models explicit spatial relationships among objects and supports memory accumulation and dynamic updates.}
    \label{fig:intro}
\end{figure}

Nonetheless, current zero-shot goal navigation approaches lack the ability to construct and retain task-relevant cognitive information across the entire scene, which limits their effectiveness in large-scale and complex environments.
First, graph-based approaches~\cite{yin2024sgnav,yin2025unigoal} rely on graph-based scene abstraction, where nodes represent objects and edges encode spatial relations, as shown in Fig.~\ref{fig:intro}~(b). However, spatial relationships are highly sensitive to viewpoint changes, which may lead to unstable graph matching and reduced robustness. 
Map-based methods~\cite{yu2023l3mvn,zhang2024trihelper,kuang2024openfmnav} maintain a semantic occupancy map of the scene, as shown in Fig.~\ref{fig:intro}~(c). While such maps provide global coverage, they are inherently large and complex, making it difficult for LLMs to directly utilize the essential information.
Moreover, both paradigms largely overlook fine-grained intrinsic and extrinsic attributes, such as object texture or background style, which hinders accurate target identification, especially in instance-level navigation tasks.
Secondly, current methods~\cite{yin2024sgnav,yu2023l3mvn} lack memory and backtracking capabilities. Once a region has been explored without detecting the goal, the agent rarely revisits it; similarly, a detected target may be missed due to viewpoint-induced matching failures. These limitations are particularly detrimental for accurate instance-level navigation.
To address these issues, a memory-aware cognitive map~(Fig.~\ref{fig:intro}~(d)) that is both information-rich and easily queryable is needed.

In this paper, we propose \textbf{\netName{}} to address the challenges discussed above. Unlike previous approaches that rely on map-based or graph-based methods to represent the environment, we additionally incorporate a sparse cognitive map, as shown in Fig.~\ref{fig:intro}, to record relational information of the scene, which is easily queryable and can be dynamically updated.
Specifically, inspired by human search behavior, we extract not only the target itself but also objects that are likely to co-occur with the goal, as these contextual objects help determine the region in which the target resides. For the IIN and TN tasks, we leverage a VLM to extract relevant objects from images or textual descriptions. In contrast, for the ON task, we rely on the commonsense reasoning capabilities of LLMs to infer objects that are likely to co-occur with the specified goal.
During exploration, we fuse multi-frame observations to enhance the accuracy of object detection and continuously update the cognitive map with all relevant object information as the agent navigates the environment.
At each exploration strategy selection stage, we introduce two distinct memory-aware exploration strategies. The goal re-validation strategy addresses failures in goal matching caused by variations in observation viewpoints, while the missing goal re-exploration strategy aims to ensure that no potential targets are overlooked in previously explored regions.
In addition, we introduce a blacklist mechanism for the cognitive map and a double-check strategy that leverages both the VLM and the cognitive map to enhance goal verification, thereby improving both efficiency and robustness.

Our contributions can be summarized as follows:
{\bf (i)}~We introduce a cognitive map that dynamically tracks objects of interest in the environment, serving as a foundation for memory-aware reasoning and exploration.
{\bf (ii)}~We design two memory-aware exploration strategies to handle missed detections and failed matches, allowing the agent to revisit potential goal locations. We also incorporate a blacklist and a VLM-based double check mechanism to enhance the efficiency of goal search and the robustness of the base navigation system.
{\bf (iii)}~Extensive evaluations on various goal-oriented navigation tasks consistently show that our approach achieves superior performance, validating its effectiveness.

\section{Related Work}
\subsection{Zero-shot navigation}
Conventional navigation methods~\cite{kwon2023renderable,ramakrishnan2022poni,chaplot2020object} require extensive training in simulated environments, which limits their generalization ability.
Recently, with the emergence of LLMs~\cite{liu2023llava,achiam2023gpt,bai2025qwen2} demonstrating strong understanding and reasoning capabilities, some works~\cite{chen2023not,kuang2024openfmnav,zhang2024trihelper} have adopted zero-shot navigation approaches, achieving competitive performance without additional training. 
According to the goal type, zero-shot navigation can be broadly categorized into three classes: ON~\cite{cai2024bridging,zhang2025multi,nie2025wmnav}, IIN~\cite{krantz2023navigating} and TN~\cite{long2024instructnav}. VLFM~\cite{yokoyama2024vlfm} use frontier-based exploration~(FBE) and use LLM for goal location reasoning. TopV-Nav~\cite{zhong2024topv} directly reasons on the top-view map with sufficient spatial information. CogNav~\cite{cao2025cognav} uses cognitive process modeling with LLMs to search for the goal object. WMNav~\cite{nie2025wmnav} integrates a world model with VLM to address ON tasks. For more challenging instance-level navigation such as IIN and TN, Mod-IIN~\cite{krantz2023navigating} leverages FBE with keypoint matching, while PSL~\cite{sun2024prioritized} and GOAT~\cite{chang2023goat} rely on additional task-specific training. UniGoal~\cite{yin2025unigoal} and InstructNav~\cite{long2024instructnav} introduce goal-oriented and language-conditioned navigation tasks~\cite{achiam2023gpt,long2024discuss,chen2024mapgpt}, respectively.
Building on these works, we incorporate spatial cognition to enable the agent to leverage memory of previously overlooked potential targets, which can benefit both language and vision-conditioned goal-oriented navigation tasks.

\subsection{Scene exploration strategy}
To enable more effective exploration and comprehensive scene understanding, various scene representation methods have been proposed, which can be broadly categorized into graph-based~\cite{liu2023bird,rajvanshi2024saynav,wu2024voronav} and map-based methods~\cite{yokoyama2024vlfm,apexnav,yu2023l3mvn}.
Graph-based representations explicitly model objects and their relationships, enabling structured reasoning and efficient relational inference for navigation.
VoroNav~\cite{wu2024voronav} introduces the Reduced Voronoi Graph to extract exploratory paths and planning nodes for efficient navigation.
UniGoal~\cite{yin2025unigoal} is a representative graph-based approach, modeling both the goal graph and the scene graph with LLM, and exploring the environment via graph matching.
TopoNav~\cite{liu2025toponav} leverages topological structures as a form of spatial memory that captures scene connectivity, adjacency, and semantic information.
Map-based representations provide a dense spatial layout of the environment, allowing the agent to plan paths, track explored regions.
L3mvn~\cite{yu2023l3mvn} first extracts the frontiers and computes their cost scores, which are then combined with LLM-based evaluations to select the optimal exploration direction.
TriHelper~\cite{zhang2024trihelper} extends L3mvn's framework with three distinct helper designs, enhancing the collision exploration and detection modules.
ApexNav~\cite{apexnav} builds on VLFM~\cite{yokoyama2024vlfm} by incorporating a semantic scoring graph and a dynamic exploration strategy, achieving efficient and reliable scene coverage.
Unlike these methods, we dynamically construct a sparse cognitive map during exploration, leveraging goal relocation and context-guided reasoning to assist target search.

\section{Problem Formulation}
We tackle goal-oriented navigation tasks, where a mobile agent is required to navigate to a specified goal $g$ in an unknown environment. Specifically, in ON, the goal is an object category~(e.g.,~``bed''); in IIN, the goal is provided as a reference image of the target object; and in TN, the goal is described through natural language instructions specifying the target.
At the beginning of each episode, the agent is initialized with a random initial pose and equipped with an egocentric RGB-D camera, along with an odometry sensor that provides its displacement and orientation relative to the initial position.
The agent executes an action $a \in \mathcal{A}$ at each time step receiving a new pose and a RGB-D observation. $\mathcal{A}$ is action space, which consists of \texttt{move\_forward,~turn\_left,~turn\_right} and \texttt{stop}.
The objective is to reach the specified goal while minimizing the path length and ensuring the agent arrives within a predefined success threshold.

\section{Method}
\subsection{Overview}
\begin{figure*}[!t]
    \centering
    \includegraphics[width=\linewidth]{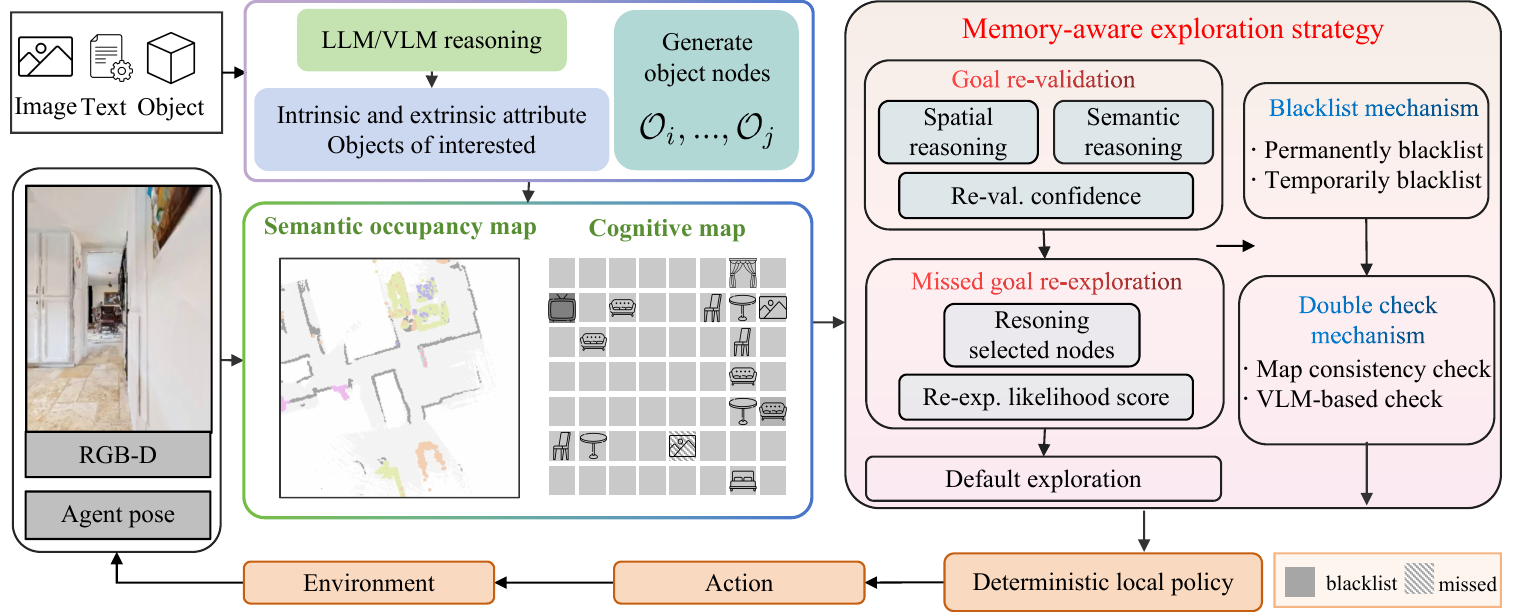}
    \caption{An overview of the \netName{} framework. 
    Our model first processes different task goals using an LLM/VLM to extract objects of interest and goal properties. During navigation, the agent iteratively performs a memory-aware exploration strategy using the cognitive map. 
    At each step, potential targets retrieved from the map are treated as temporary goals and verified by a VLM-based double-check mechanism. 
    We further maintain a blacklist to record failed goals, preventing repeated exploration of unpromising locations.
}
    \label{fig:pipeline}
\end{figure*}

The overview of \netName{} is shown in Fig.~\ref{fig:pipeline}. We first introduce our base navigation system, which is suitable for goal-oriented navigation tasks~(Sec.~\ref{IV-B}). 
During exploration, the agent performs multi-frame fusion to accurately detect objects of interest and constructs a sparse cognitive map, which serves as its primary form of spatial and semantic memory~(Sec.~\ref{IV-C}). 
Next, we propose a memory-aware exploration strategy that queries the cognitive map to identify potential goals in explored regions, thus correcting for earlier detection and matching errors~(Sec.~\ref{IV-D}). 
To further ensure robustness and efficiency, we introduce a blacklist mechanism within the cognitive map and a VLM-based double check strategy for goal verification~(Sec.~\ref{IV-E}).
We also provide the detailed algorithmic pipeline of \netName{}, as shown in Algorithm~\ref{algo:memory_exploration}.

\subsection{Base navigation system}\label{IV-B}
Our base navigation system employs a frontier-based exploration strategy, which incrementally builds a semantic occupancy map~$\mathcal{M}_{occ}$ to guide navigation towards a specific goal.

\noindent\textbf{Semantic occupancy map.} At each time step $t$, the agent employs Mask R-CNN~\cite{he2017mask} for object segmentation and performs inverse perspective projection using the egocentric depth and pose to construct a semantic occupancy map $\mathcal{M}_{\text{occ}} \in \mathbb{R}^{M \times M \times C}$. The map represents free, occupied, and unknown regions, where occupied cells are annotated with semantic labels of key objects of interest. Here, $C$ denotes the number of semantic classes per grid cell, which is set to $8$ in this work.
Frontiers are then defined as free cells adjacent to at least one unknown cell, serving as candidates for exploration and path planning. Following~\cite{apexnav}, we incrementally update frontiers and group them into clusters, which are then approximated by their centroids using Principal Component Analysis~(PCA) to improve computational efficiency.

\noindent\textbf{Exploration mechanism.}
For exploration stability and continuity, we follow prior work~\cite{yin2025unigoal} and adopt a long-term exploration mechanism. This long-term goal is processed by a deterministic local policy~\cite{sethian1999fast} to obtain actions.
During navigation, the agent operates in a continuous perception-exploration loop. It periodically sets the nearest frontier as a long-term target every 20 time steps, while simultaneously employing Mask R-CNN~\cite{he2017mask} to detect candidate objects. If an object matching the target category is found, the agent overrides its current exploration plan, designates the object as a temporary goal and navigates directly towards it for verification. For IIN, verification is performed via LightGlue~\cite{lindenberger2023lightglue} key point matching, whereas for ON and TN, the decision relies solely on category-level matching.

\begin{figure}[t]
    \centering
    \includegraphics[width=\linewidth]{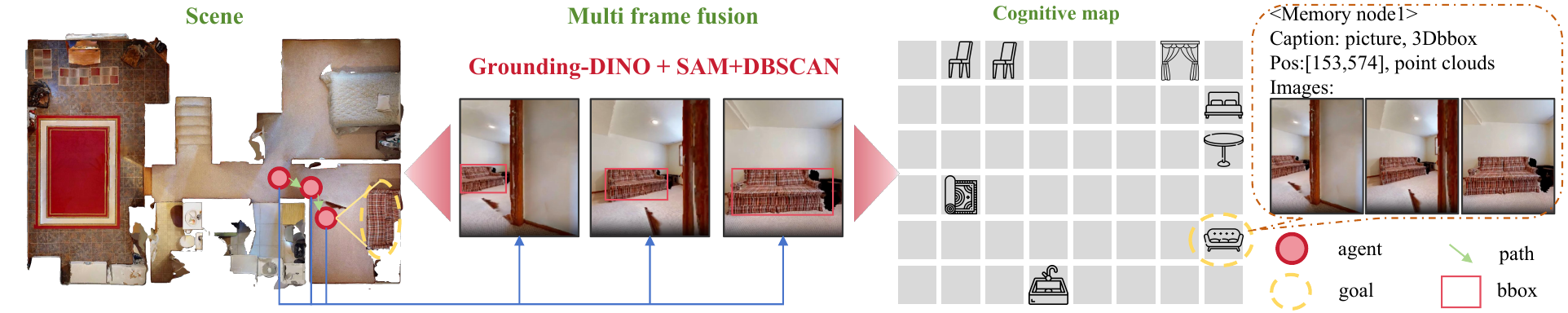}
    \caption{In our cognitive map construction process, the agent observes multiple frames and, after performing detection and segmentation fusion, stores the results into the cognitive map.}
    \label{fig:meth_cog}
\end{figure}

\subsection{Dynamic cognitive map}\label{IV-C}
To effectively ground a LLM for navigation, an ideal scene representation must be rich in both semantic and spatial information, while also being efficiently queryable. While existing methods are powerful, they often struggle to meet all these criteria simultaneously. Graph-based methods~\cite{yin2024sgnav,yin2025unigoal}, for instance, excel at capturing spatial relationships but typically lack the detailed semantic attributes required for complex reasoning. Conversely, semantic occupancy maps~\cite{zhong2024topv,kuang2024openfmnav} store comprehensive information, but their dense structure is inherently inefficient to query. This makes it difficult to provide an LLM with the precise, task-relevant data needed to fully unlock its reasoning capabilities. To overcome these limitations, we introduce a sparse cognitive map, as shown in Fig.~\ref{fig:meth_cog}, that maintains both semantic and spatial information of key objects and is designed to unlock the full reasoning capabilities of the LLM.

Our cognitive map focuses on objects of interest that are most likely to influence the agent's actions during navigation. To this end, we define a base set of detection categories 
$\mathcal{C}_{\text{base}} = \{\texttt{chair}, \texttt{plant}, \texttt{sofa}, \texttt{bed}, \texttt{toilet}, \texttt{tv}, \texttt{table}, \dots \}$,
which includes commonly encountered objects that may affect planning and decision-making. Beyond these predefined categories, we leverage the reasoning capability of the VLM~\cite{bai2025qwen2} to dynamically extend the set of objects relevant to the current task. For ON, the LLM predicts potential co-occurring objects based on commonsense knowledge of the target; for more challenging task IIN and TN, the VLM analyzes the given goal image and textual description to extract additional objects surrounding the goal. The inferred objects $\mathcal{C}_{\text{LLM}}$ are then merged with $\mathcal{C}_{\text{base}}$ to form a task-adaptive candidate set: $\mathcal{C}_{\text{cog}} = \mathcal{C}_{\text{base}} \cup \mathcal{C}_{\text{LLM}}$.
This process enables the cognitive map to maintain a comprehensive and task-aware representation of the environment, providing precise and relevant information to the downstream VLM for reasoning and decision making.

Inspired by~\cite{yin2024sgnav}, we adopt a lightweight multi-frame fusion detection strategy to construct and maintain the cognitive map during navigation. At each time step $t$, we leverage Grounding-DINO~\cite{liu2024grounding} and SAM~\cite{sam} to perform object detection and segmentation on the current RGB image $I_t$. For each detected object, we construct its corresponding point cloud~$P_t^i$ with the depth map by projecting the segmented region into 3D space. To improve efficiency and reduce computational overhead, we apply a voxel downsampling operation to each~$P_t^i$ and further remove outliers using DBSCAN. 
Subsequently, a fusion step is performed to maintain a consistent and robust representation of the environment, we fuse the current frame's object candidates with the previously accumulated detections based on the Intersection-over-Union of their point clouds. 

For each candidate cluster, once an object has been detected more than three times, it is considered a confirmed detection and inserted into $\mathcal{M}_{\text{cog}}$. Each confirmed object is represented as an object node: $\mathcal{O}_j = \langle c_j, P_j, (x_j, y_j), \mathcal{I}_j \rangle$, where $c_j$ denotes the object category, $P_j$ is the aggregated object-level point clouds, $(x_j, y_j)$ is the estimated 2D center coordinate, and $\mathcal{I}_j$ is a ranked list of the original, full-frame observation images associated with the object. The ranking of $\mathcal{I}_j$ is determined by a quality score, calculated as the weighted sum of the detection confidence and the object's relative area within the image frame:
$
\text{score} = 0.5 \times \text{confidence} + 0.5 \times \frac{\text{Area}_{\text{object}}}{\text{Area}_{\text{image}}}
$
The list is then sorted in descending order to retain the most informative views.
Consequently, the cognitive map at time $t$ can be represented as a set of object nodes:
$\mathcal{M}_{\text{cog}}^t = \{ \mathcal{O}_1, \mathcal{O}_2, \dots, \mathcal{O}_N \}$,
where each $\mathcal{O}_j$ is defined as above and $N$ denotes the total number of confirmed objects in the map.

\subsection{Memory-aware exploration strategy}\label{IV-D}

The robustness of current instance-level navigation tasks like IIN and TN is often compromised by two key challenges in complex environments. The first is verification failure, where low-level feature matching~\cite{chang2023goat} fails to confirm a detected object's identity across severe viewpoint changes. The second is perception failure, where the target is never detected at all due to occlusion or poor lines of sight. These issues highlight the need for a higher-level reasoning mechanism that integrates semantic knowledge, spatial relationships, and contextual cues to guide exploration more intelligently.

Our approach tackles these challenges by creating a synergy between our $\mathcal{M}_{\text{cog}}$ and VLM. The VLM leverages the rich semantic and spatial information in the map to infer likely goal locations. Building on this foundation, we introduce two specific memory-aware strategies: goal re-validation and missed goal re-exploration, shown in Algorithm~\ref{algo:memory_exploration}.
For consistency across different navigation tasks, we define a unified goal representation as $G = \{c, a_{\text{int}}, a_{\text{ext}}\}$, where $c$ denotes the category, while $a_{\text{int}}$ and $a_{\text{ext}}$ correspond to intrinsic attributes~(e.g., texture, shape) and extrinsic attributes~(e.g., surrounding context), respectively. 
For IIN, all three components are extracted from the instance image via the VLM; for TN, they are derived directly from the textual description; and for ON, only the category $c$ is available, with the other components set to None. We elaborate more on this in the supplementary.

\noindent\textbf{Goal re-validation.}
Our goal re-validation strategy addresses the challenge of previously missed targets by re-assessing candidate objects in the cognitive map using spatial and semantic reasoning. The process begins by selecting all object nodes $\mathcal{O}^c$ that match the goal category~$c_{\text{goal}}$. Then, for each candidate $\mathcal{O}_j \in \mathcal{O}^c$, we evaluate its spatial consistency. This is achieved by first collecting its neighboring objects $\mathcal{N}_j = \{ c_k \mid \| (x_k,y_k)-(x_j,y_j) \| \le r \}$, where $r$ denotes a predefined spatial radius,
and computing a contextual score~$s^{\text{spa}}_j$, based on the overlap between their categories and the LLM-inferred context set~$\mathcal{C}_{\text{LLM}}$:
\begin{equation}
s^{\text{spa}}_j = 
\frac{|\{ c_k \in \mathcal{N}_j \} \cap \mathcal{C}_{\text{LLM}}|}{|\{ c_k \in \mathcal{N}_j \} \cup \mathcal{C}_{\text{LLM}}|}.
\end{equation}
For semantic reasoning, we select the highest scoring image $I_j^*$ from $I_j$ and feed it into the VLM, together with a prompt that evaluates the similarity between the candidate and the goal based on intrinsic and extrinsic attributes:
\begin{equation}
s = \text{VLM}\big(I_j^*,\ \text{Prompt}_{\text{rev}}, G\big).
\end{equation}
The $s$ contains two scores: $s^{\text{int}}_j \in [0,1]$ for semantic~(visual appearance) similarity and $s^{\text{ext}}_j \in [0,1]$ for spatial~(surrounding context) similarity. The VLM is instructed with the following concise prompt:

\textit{``Given a goal with intrinsic and extrinsic attributes, assess a candidate object across multiple images by separately evaluating intrinsic and extrinsic consistency and providing a brief reasoning.''}

We then combine these spatial and semantic signals into a unified relocation confidence:
\begin{equation}
\ell_{\text{rev}}(j) = (s^{\text{int}}_j + s^{\text{ext}}_j + s^{\text{spa}}_j)/3
\end{equation}
If $\ell_{\text{rev}}(j)$ exceeds a threshold $\tau_{\text{rev}}$, the agent navigates back to $(x_j,y_j)$, performs a 360-degree scan to gather new viewpoints, and then attempts to rematch the target; otherwise, it continues missed goal re-exploration.

\noindent\textbf{Missed goal re-exploration.}
This strategy is designed for situations where the agent has explored a region but has not detected the goal. In such cases, it is often possible to observe surrounding objects that are highly correlated with the goal, providing useful cues for inferring whether the goal might be located nearby.
The re-exploration process begins by selecting all object nodes from the cognitive map, $\mathcal{O}_{\text{sel}} = \{\, \mathcal{O}_j \in \mathcal{M}_{\text{cog}}^t \mid c_j \in \mathcal{C}_{\text{LLM}}\}$. For each selected object $\mathcal{O}_j$, we leverage the VLM to infer the likelihood that the goal is hidden in its vicinity. To achieve this, we provide the VLM with the diverse perspectives stored in the object's image list $I_j$ and a carefully designed prompt. The VLM then outputs a single likelihood score~ $\ell_{\text{ree}}(j)$:
\begin{equation}
\ell_{\text{ree}}(j) = \text{VLM}(I_j, \text{prompt}_{\text{ree}}, G).
\end{equation}
The prompt is designed to be concise while capturing both semantic and spatial reasoning:

\textit{``Based on the visual evidence and the goal's category and extrinsic attributes, infer the likelihood that the target is nearby but out of view.''}

The agent's next action is determined by this score. We define the next target location~$\mathbf{pos}^*=(x_n,y_n)$, based on a threshold, $\tau_{\text{ree}}$:
\begin{equation}
\mathbf{pos}^* =
\begin{cases}
\mathbf{pos}_j, & \text{if } \ell_{\text{ree}}(j) > \tau_{\text{ree}} \\
\mathbf{pos}_{\text{fro}}, & \text{otherwise}
\end{cases}
\end{equation}
where $\mathbf{pos}_{\text{fro}}$ is the next target selected by the default exploration strategy. If the likelihood exceeds the threshold, the agent returns to the vicinity of the correlated object and performs a 360-degree scan of the area to search for the potentially occluded goal. Otherwise, it continues its default exploration.

\subsection{Mechanisms for maintaining exploration stability}
\label{IV-E}

\IncMargin{0.5em}
\DontPrintSemicolon
\begin{algorithm}[t]
\caption{Pipeline of \netName{}}
\label{algo:memory_exploration}
\SetKwInOut{AlgoInput}{Input/Output}
\SetKwFunction{Observation}{Observation}
\SetKwFunction{DynamicCognitiveMap}{DynamicCognitiveMap}
\SetKwFunction{VLM}{VLM}
\SetKwFunction{LLM}{LLM}
\SetKwFunction{FMM}{FastMarching}

\SetCommentSty{\color{gray}\scriptsize}
\SetKwComment{Comment}{}{}

\definecolor{lightblue}{RGB}{69, 110, 196}
\definecolor{lightgray}{RGB}{155, 155, 155}
\newcommand{\RightComment}[1]{{\hfill{\color{lightgray}$\triangleright$~#1}}}
\newcommand{\LeftComment}[1]{{{\color{lightgray}$\triangleright$~#1}}}

\AlgoInput{Target category $c$, Blacklist $\mathcal{B}_0 \gets \emptyset$; 
Task completion status}
\Repeat{$a_t$ is \text{stop}}{
    $I_t \gets$ \Observation{$t$},
    $\mathcal{M}_{\text{occ}}^t \gets$ \DynamicCognitiveMap~($I_t$)\;
    \LeftComment{Memory-aware exploration strategy}\;
    $goal_t \gets
    \begin{cases}
    \text{Goal re-validation}, & \text{if } c \in \mathcal{M}_{\text{occ}}^t \text{ but fails to match}\\
    \text{Missed goal re-exploration}, & \text{if related objects found}\\
    \text{Default exploration}, & \text{otherwise}
    \end{cases}$
    
    \LeftComment{Redecision \& Refinement Phase}\;
    \If{$goal_t$ is temporary goal}{
        \lIf{matching fails}{$\mathcal{B}_{t+1} \gets \mathcal{B}_t \cup \{goal_t\}$; \textbf{continue}}
        \lElse{$goal_t \gets \textit{DoubleCheck}(goal_t, c)$}
    }
    \LeftComment{Execution via Deterministic local policy}\;
    $a_t \gets \FMM{$goal_t$}$\;
}
\end{algorithm}
To ensure robust and stable exploration, we introduce two mechanisms designed to prevent repeated failures and mitigate confirmation errors: a blacklist for failed targets and a VLM-based double-check for potential targets.

\noindent\textbf{Blacklist mechanism.}
To prevent the agent from repeatedly pursuing confirmed non-targets, we maintain a blacklist~$\mathcal{B}$, within the cognitive map. An object node is added to $\mathcal{B}$ under two conditions: (i)~if it fails VLM-based re-validation three consecutive times, it is temporarily blacklisted; (ii)~if it passes re-validation from a distance but fails the final verification upon arrival, it is permanently blacklisted. Temporarily blacklisted objects can be removed if significant new visual evidence becomes available, whereas permanently blacklisted objects are perpetually ignored as goal candidates.

\noindent\textbf{Double check mechanism.}
To mitigate errors from false positives or temporary detection loss, we employ a double check mechanism before the agent commits to a temporary goal. The implementation of this mechanism is tailored to the type of task. For ON, we perform a map consistency check that leverages the cognitive map's memory. When a candidate object is detected, we identify its region on $\mathcal{M}_{\text{occ}}^t$ and then query our $\mathcal{M}_{\text{cog}}^t$ for any previously confirmed object nodes within this area. The goal is confirmed if the category of any retrieved node matches $c_{\text{goal}}$. For the more demanding IIN and TN tasks, which require precise instance matching, we employ a VLM-based double-check. In this case, the VLM directly compares the current visual observation~$I_t$ with the detailed goal specification~$G$ to infer if the specific target instance is truly present.
\section{Experiments}

\subsection{Experimental setup}
\textbf{Datasets}:
We evaluate our \netName{} in Habitat~\cite{szot2021habitat} simulator on two large-scale datasets: \textit{HM3Dv2}(HM3D-Semantics-v0.2 from 2023 Habitat Challenge, 1000~episodes, 36~scenes, 6~goal categories) for instance-level task, IIN and TN.
\textit{HM3Dv1}~(HM3D-Semantics-v0.1 from 2022 Habitat Challenge, 2000~episodes, 20~scenes, 6~goal categories) for category-level task, ON. 

\noindent\textbf{Evaluation metrics}:
We report Success Rate~(SR) and Success weighted by inverse Path Length~(SPL). SR denotes the proportion of successful navigation episodes, while SPL further accounts for efficiency by comparing the executed path length with the optimal one~(SPL is 0 for failures).

\noindent\textbf{Implementation details}:
The agent moves $0.25$\,m per step and rotates in increments of $30^{\circ}$.  
An onboard camera, mounted at a height of $0.88$\,m, provides $640 \times 480$ RGB-D observations.  
For IIN and TN tasks, the perception range is set to $[0, 30]$\,m, and each episode is capped at $1000$ steps with a success threshold of $1.0$\,m. 
For ON tasks, the perception range is restricted to $[0, 5]$\,m, with a maximum of $500$ steps with a success threshold of $0.2$\,m.  
For the semantic occupancy map, we set the size to $720 \times 720$ with a resolution of $0.05$\,m, corresponding to a large-scale area of $36$\,m $\times$ $36$\,m. We set $\tau_{\text{rev}}$ and $\tau_{t}$ as 0.7 and $\tau_{\text{ree}}$ as 0.5.
We use Qwen2.5-VL-7B as our LLM and VLM. All experiments are conducted on two NVIDIA GeForce RTX 4090 GPUs.
Additional experimental details and more experiments, will be provided in the supplementary material.

\noindent\textbf{Compared methods}: 
We evaluate \netName{} against state-of-the-art baselines across three navigation tasks. 
For IIN and TN, which involve fine-grained goal specifications, we compare with supervised methods including Krantz et al.~\cite{krantz2022instance}, OVRL-v2-IIN~\cite{yadav2023ovrl}, IEVE~\cite{lei2024instance}, PSL~\cite{sun2024prioritized}, and GOAT~\cite{chang2023goat}, as well as zero-shot approaches such as Mod-IIN~\cite{krantz2023navigating} and UniGoal~\cite{yin2025unigoal}. 
Since dedicated zero-shot methods for TN remain limited, we also report results of these approaches across tasks for completeness. 
We reproduce UniGoal using the official code with the Qwen2.5-VL-7B~\cite{bai2025qwen2} backbone, which is available for IIN and TN.
For ON, we compare against supervised methods ZSON~\cite{majumdar2022zson} and CoW~\cite{gadre2023cows}, as well as zero-shot methods ESC~\cite{zhou2023esc}, OpenFMNav~\cite{kuang2024openfmnav}, L3MVN~\cite{yu2023l3mvn}, VLFM~\cite{yokoyama2024vlfm}, SG-Nav~\cite{yin2024sgnav}, and WMNav~\cite{nie2025wmnav}. 
For WMNav, we report results under different VLM backbones, including Qwen2.5-VL-7B and Gemini-1.5Pro~\cite{team2024gemini}.


\begin{table*}[t!]
    \centering
    \caption{Results of instance-image-goal navigation, text-goal navigation and object-goal navigation on HM3D. We report SR and SPL for state-of-the-art methods under different settings. Universal goal-oriented navigation methods are highlighted in gray. For UniGoal~\cite{yin2025unigoal}, we reproduce its results using the official open-source code, which is only available for IIN and TN.}
    \label{tab:main}
    \begin{tabular}{lcccccccc}
        \toprule
        \multirow{2}*{\textbf{Method}} & \multirow{2}*{\textbf{Zero-shot}} &
        \multirow{2}*{\textbf{Universal}} &
        \multicolumn{2}{c}{\textbf{InsINav}} & \multicolumn{2}{c}{\textbf{TextNav}} & \multicolumn{2}{c}{\textbf{ObjNav}}\\
        \cmidrule(lr){4-5} \cmidrule(lr){6-7} \cmidrule(lr){8-9}
        & & &SR$\uparrow$& SPL$\uparrow$ & SR$\uparrow$ & SPL$\uparrow$ & SR$\uparrow$ & SPL$\uparrow$ \\
        \midrule
        ZSON~\cite{majumdar2022zson} & $\times$& $\times$  & -- & -- & -- & -- & 25.5 & 12.6  \\
        Cow~\cite{gadre2023cows} & $\times$ & $\times$& -- & -- & -- & -- & 32.0 & 18.1 \\
        Krantz et al.~\cite{krantz2022instance}  & $\times$ & $\times$& 8.3 & 3.5 & -- & -- & -- & --\\
        OVRL-v2-IIN~\cite{yadav2023ovrl} & $\times$ & $\times$ & 24.8 & 11.8 & -- & -- & -- & -- \\
        IEVE~\cite{lei2024instance} & $\times$ & $\times$& 70.2 & 25.2 & -- & -- & -- & -- \\
        \rowcolor{mygray}PSL~\cite{sun2024prioritized} & $\times$  & $\checkmark$ & 23.0 & 11.4 & 16.5 & 7.5 & 42.4 & 19.2\\
        \rowcolor{mygray}GOAT~\cite{chang2023goat} & $\times$& $\checkmark$ & 37.4 & 16.1 & 17.0 & 8.8 & 50.6 & 24.1 \\
        \midrule
        ESC~\cite{zhou2023esc} & $\checkmark$ & $\times$ & -- & -- & -- & -- & 39.2 & 22.3\\
        OpenFMNav~\cite{kuang2024openfmnav} & $\checkmark$ & $\times$ & -- & -- & -- & -- & 52.5 & 24.1\\
        L3MVN~\cite{yu2023l3mvn} & $\checkmark$ & $\times$ & -- & -- & -- & -- & 50.4 & 23.1 \\
        VLFM~\cite{yokoyama2024vlfm} & $\checkmark$ & $\times$ & -- & -- & -- & -- & 52.4 & 30.3 \\
        SG-Nav~\cite{yin2024sgnav} & $\checkmark$ & $\times$& -- & -- & -- & -- & 54.0 & 24.9 \\
        WMNav\textsubscript{Gemini}~\cite{nie2025wmnav} & $\checkmark$ & $\times$& -- & -- & -- & -- & \textbf{58.1} & \textbf{31.2} \\
        WMNav\textsubscript{Qwen2.5}~\cite{nie2025wmnav} & $\checkmark$ & $\times$& -- & -- & -- & -- & 46.1 & 20.7 \\
        Mod-IIN~\cite{krantz2023navigating} & $\checkmark$ & $\times$  & 56.1 & 23.3 & -- & -- & -- & --\\
        \rowcolor{mygray}UniGoal~\cite{yin2025unigoal} & $\checkmark$ & $\checkmark$& 60.2 & 23.7 & 20.2 & 11.4 & 54.5 & 25.1 \\
        \rowcolor{mygray}UniGoal\textsubscript{Qwen2.5}~\cite{yin2025unigoal}\textsuperscript{†} & $\checkmark$& $\checkmark$  & 58.5 & 21.9 & 18.8 & 9.2 & -- & -- \\
        \rowcolor{mygray}\netName{}~(Ours)& $\checkmark$ & $\checkmark$& \textbf{61.4} & \textbf{23.9} & \textbf{22.9} & \textbf{11.7} & 54.8 & 25.1 \\
        \bottomrule
    \end{tabular}
    
\end{table*}

\subsection{Comparison with state-of-art methods}
We compare our \netName{} with state-of-the-art goal-oriented navigation methods under different settings, including supervised, zero-shot, and universal approaches, across the three studied tasks, as shown in Table~\ref{tab:main}.
For complex instance-level navigation tasks~(IIN and TN), our method achieves clear improvements over prior universal approaches such as UniGoal and GOAT. 
Specifically, on IIN, our \netName{} surpasses the zero-shot methods Mod-IIN and UniGoal by 5.3\% and 1.2\% in SR, respectively. When compared with PSL and GOAT, our method achieves significant improvements in SR and SPL. 
For TN, which is inherently more challenging as it requires navigating objects solely based on textual descriptions, \netName{} outperforms the supervised methods PSL and GOAT by 6.4\% and 5.9\% in SR, respectively, while also surpassing UniGoal by 2.7\% in SR and 0.3 in SPL. 
The improvements on the IIN and TN tasks demonstrate that our method can effectively enhance navigation success rates by searching for memory targets of interest within the cognitive map and re-exploring them.

For the ON task, \netName{}~achieves a 0.3\% improvement in SR over UniGoal. 
We claim that this relatively modest gain is due to the nature of ON, where locating any instance of the target category is sufficient, thereby reducing the reliance on precise instance-level discrimination and memory-aware cognitive reasoning.
Compared with WMNav, which benefits from panoramic observations and a strong VLM backbone, our method remains competitive. Notably, when both approaches employ the same VLM~(Qwen2.5-VL-7B), \netName{} surpasses WMNav by 8.7\% in SR and 4.4\% in SPL, demonstrating the effectiveness of our memory-aware re-exploration strategy under a controlled backbone setting.


\begin{table}[t]
    \centering
    \caption{Effect of each submodule on HM3D~(IIN) benchmark. Abbreviations: CMap = cognitive map, GR = goal re-validation, MGR = missed goal re-exploration, BL = blacklist mechanism, DC = double check mechanism.
}
    \vspace{-5pt}
    \setlength\tabcolsep{5pt}
    \begin{tabular}{cccccccc}
        \toprule
        Exp&CMap & GR & MGR & BL& DC &SR$\uparrow$ & SPL$\uparrow$ \\
        \midrule
        a& & & & & &55.8 & 21.0 \\
        b& \checkmark& \checkmark& & \checkmark& &57.4 & 21.6 \\
        c& \checkmark& \checkmark& \checkmark& \checkmark& &60.1 & 22.6 \\
        d&\checkmark & \checkmark & \checkmark& & &52.1 & 18.5 \\
        e& \checkmark& & & & \checkmark&57.8 & 22.0 \\
        f&\checkmark & \checkmark & \checkmark & \checkmark& \checkmark&\textbf{61.4} & \textbf{23.9} \\
        \bottomrule
    \end{tabular}
    \label{tab:maes}
\end{table}

\subsection{Ablation study}
We conducted experiments on the HM3D dataset to validate the effectiveness of each proposed module and reported the final results on the IIN task. 

\noindent\textbf{Main ablation study}
We conduct a comprehensive ablation study, presented in Table \ref{tab:maes}, to validate the effectiveness of each proposed module.
Our analysis begins with the base navigation system~(Exp a), which achieves an SR of 55.8\% and an SPL of 21.0\%. By incorporating the cognitive map, goal re-validation, and the blacklist, performance improves to 57.4\% SR~(Exp b). Further integrating the missing goal re-exploration strategy, boosts performance to 60.1\% SR~(Exp c). These results validate the effectiveness of our memory-aware exploration strategy.
We then remove the blacklist from our fully-equipped memory-aware exploration strategy~(comparing Exp c with Exp d) causes a significant performance drop from 60.1\% to 52.1\% SR. This result underscores the blacklist's critical role in preventing redundant exploration and enhancing navigation efficiency. We also analyze the positive interaction between the cognitive map and the double-check mechanism. As shown in Exp e, integrating these two modules leads to a notable performance gain over the baseline, reaching 57.8\% SR. This confirms that a persistent semantic memory and a robust verification process are highly complementary, leading to more reliable navigation.
Finally, our full model~(Exp f), which integrates all components, achieves the optimal performance with an SR of 61.4\% and an SPL of 23.9\%. These results confirm that each module provides a distinct and valuable contribution, and their powerful synergy is key to achieving robust and efficient navigation.

\begin{figure*}[t]
    \centering
    \includegraphics[width=\linewidth]{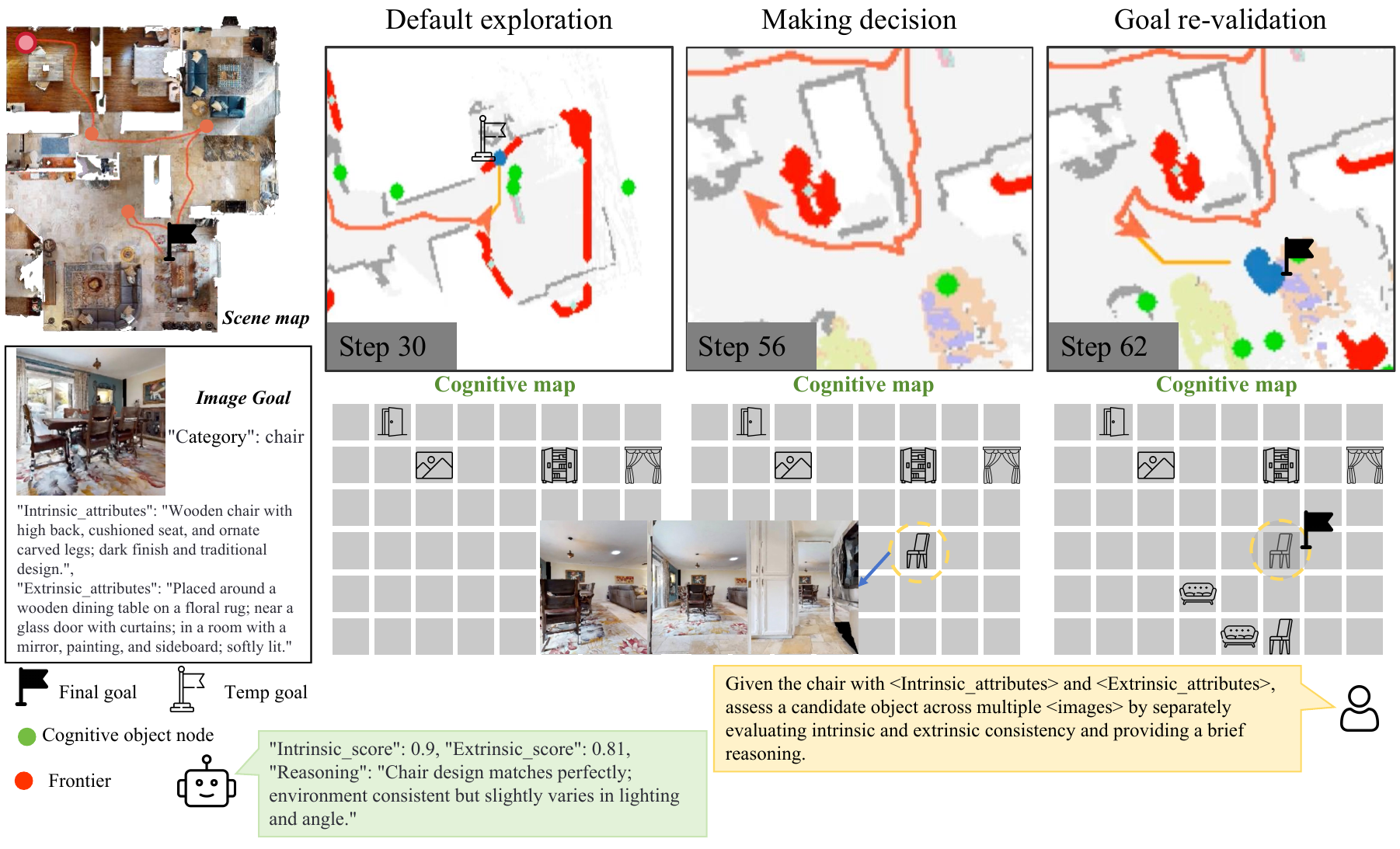}
    \caption{Demonstration of the decision process of \netName. The agent makes a new exploration decision upon reaching each temporary goal. Red dots on the occupancy map represent boundary points, while green dots indicate the projection of object nodes from the cognitive map onto the corresponding positions in the occupancy map.}
    \label{fig:vis}
\end{figure*}

\begin{table}[t]
\centering

\begin{minipage}{0.52\linewidth}
\centering
\caption{Ablation study on threshold.}
\vspace{-5pt}
\setlength\tabcolsep{4pt}
\begin{tabular}{cccc}
\toprule
Exp & Threshold & SR$\uparrow$ & SPL$\uparrow$ \\
\midrule
a   & $\tau_{rev}=0.5, \tau_{ree}=0.3$ & 54.3 & 19.6 \\
b  & $\tau_{rev}=0.6, \tau_{ree}=0.4$ & 59.1 & 21.9 \\
c & $\tau_{rev}=0.7, \tau_{ree}=0.5$ & \textbf{60.1} & \textbf{22.6} \\
d  & $\tau_{rev}=0.8, \tau_{ree}=0.6$ & 58.8 & 21.8 \\
\bottomrule
\end{tabular}
\label{tab:threshold}
\end{minipage}
\hfill
\begin{minipage}{0.46\linewidth}
\centering
\caption{Ablation study of \netName{} with different MLLMs.}
\vspace{-5pt}
\setlength\tabcolsep{2pt}
\begin{tabular}{ccc}
\toprule
MLLM & SR$\uparrow$ & SPL$\uparrow$ \\
\midrule
Llama3.2-Vision-11B~\cite{meta2024llama} & 60.2 & 20.7 \\
Llava-7B~\cite{liu2023llava} & 58.7 & 19.3 \\
Qwen2.5-VL-7B~\cite{bai2025qwen2}  & \textbf{61.4} & \textbf{23.9} \\
\bottomrule
\end{tabular}
\label{tab:mllm}
\end{minipage}

\end{table}

\noindent\textbf{Ablation study of threshold.}
We study the impact of different thresholds in the dynamic exploration strategy by varying $\tau_{rev}$ and $\tau_{ree}$, evaluated with SR and SPL on the HM3D dataset for IIN. 
As shown in Table~\ref{tab:threshold}, the best performance is achieved at $\tau_{rev}=0.7$ and $\tau_{ree}=0.5$. 
Both excessively high and low thresholds degrade performance: overly strict settings hinder effective exploration, while overly loose ones trigger frequent memory re-traversal and waste navigation steps. 
These results highlight the importance of proper threshold selection for balanced and efficient exploration.

\noindent\textbf{Ablation study of different MLLMs.}
We further compare different MLLMs on the IIN task. As shown in Table~\ref{tab:mllm}, Qwen2.5-VL-7B achieves the best performance, outperforming Llama3.2-Vision-11B and Llava-7B in both SR and SPL. This is likely due to the stronger scene understanding and reasoning capabilities of Qwen2.5-VL-7B. Ideally, our framework is model-agnostic and can naturally benefit from future advances in MLLMs capabilities.


\begin{figure*}[t]
    \centering
    \includegraphics[width=\linewidth]{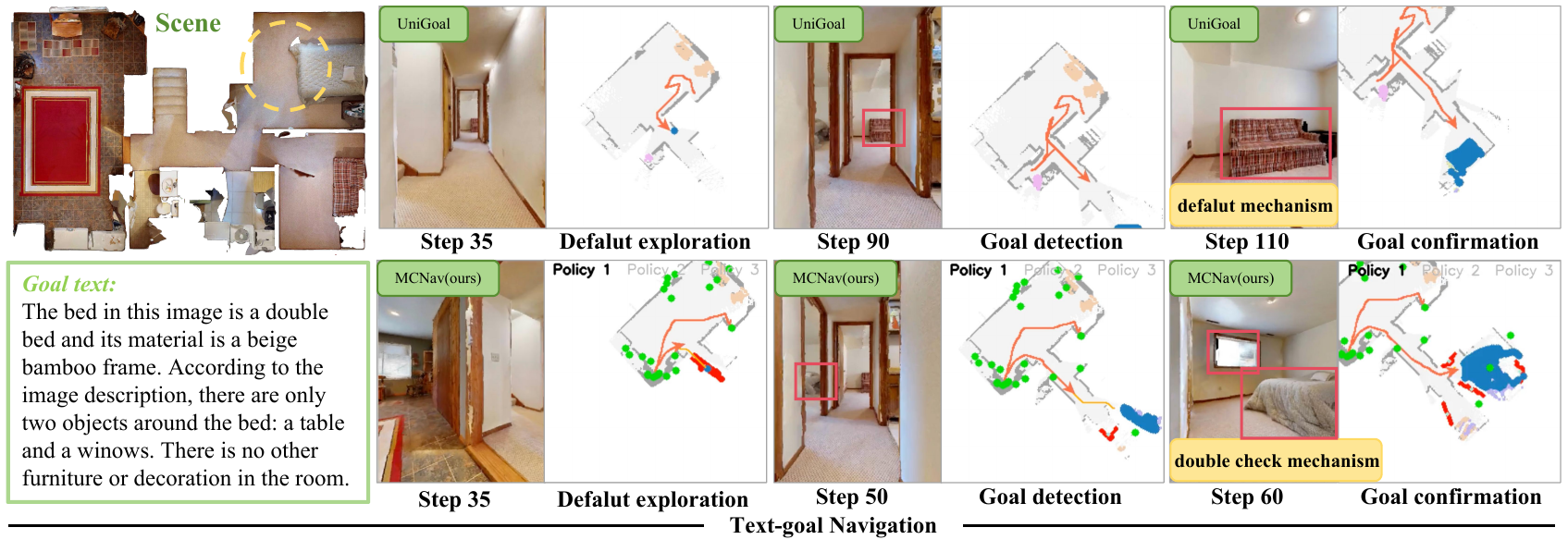}
    \caption{Comparison between UniGoal and \netName{} on TN. The double-check mechanism improves navigation reliability.}
    \label{fig:vis_dcm}
\end{figure*}



\subsection{Qualitative results}
We visualize \netName{} to illustrate its memory-aware exploration strategy. 
Using goal re-validation as an example (Fig.~\ref{fig:vis}), we show the decision process on the IIN task. 
The image goal is decomposed by the VLM into intrinsic and extrinsic attributes. 
During exploration, the agent initially fails to recognize the goal when passing it. At a later decision stage, memory recall enables re-validation, leading to successful navigation.
We further compare with UniGoal~\cite{yin2025unigoal} on challenging TN task~(Fig.~\ref{fig:vis_dcm}). 
UniGoal relies mainly on category-level graph matching, which can cause premature or incorrect goal confirmation. 
In contrast, \netName{} adopts a double-check mechanism that considers both intrinsic and extrinsic attributes, reducing misidentification and improving navigation reliability.

\section{Conclusion}

In this paper, we presented \netName{}, a unified and zero-shot framework designed to address navigation failures caused by insufficient memory and reasoning. Our approach is centered on a dynamic cognitive map that stores task-relevant information about explored areas. Based on this map, we introduced two novel memory-aware strategies powered by a VLM: goal re-validation, which re-assesses previously seen objects to correct matching failures, and missing goal re-exploration, which infers the likelihood of unseen targets from contextual cues. The framework's robustness is further enhanced by a blacklist mechanism and a double-check strategy. Extensive experiments show that \netName{} achieves state-of-the-art performance across multiple benchmarks without any task-specific training, demonstrating that integrating a cognitive memory with VLM-based reasoning is a highly effective paradigm for more intelligent navigation agents.

%
%
\bibliographystyle{splncs04}
\bibliography{main}

@String(CVPR  = {IEEE Conf. Comput. Vis. Pattern Recog.})

@String(ICCV  = {Int. Conf. Comput. Vis.})

@String(ECCV  = {Eur. Conf. Comput. Vis.})

@String(NeurIPS = {Adv. Neural Inform. Process. Syst.})

@String(ICML  = {Int. Conf. Mach. Learn.})

@String(CVPR  = {CVPR})

@String(ICCV  = {ICCV})

@String(ECCV  = {ECCV})

@String(NeurIPS = {NeurIPS})

@String(ICML  = {ICML})

@article{apexnav,
  author={Zhang, Mingjie and Du, Yuheng and Wu, Chengkai and Zhou, Jinni and Qi, Zhenchao and Ma, Jun and Zhou, Boyu},
  journal={IEEE RA-L}, 
  title={ApexNav: An Adaptive Exploration Strategy for Zero-Shot Object Navigation with Target-centric Semantic Fusion}, 
  year={2025},
  }

@inproceedings{yin2025unigoal,
  title={Unigoal: Towards universal zero-shot goal-oriented navigation},
  author={Yin, Hang and Xu, Xiuwei and Zhao, Linqing and Wang, Ziwei and Zhou, Jie and Lu, Jiwen},
  booktitle={CVPR},
  year={2025}
}

@article{yin2024sgnav,
  title={Sg-nav: Online 3d scene graph prompting for llm-based zero-shot object navigation},
  author={Yin, Hang and Xu, Xiuwei and Wu, Zhenyu and Zhou, Jie and Lu, Jiwen},
  journal={NeurIPS},
  year={2024}
}

@article{kuang2024openfmnav,
  title={Openfmnav: Towards open-set zero-shot object navigation via vision-language foundation models},
  author={Kuang, Yuxuan and Lin, Hai and Jiang, Meng},
  journal={arXiv preprint arXiv:2402.10670},
  year={2024}
}

@article{zhong2024topv,
  title={Topv-nav: Unlocking the top-view spatial reasoning potential of mllm for zero-shot object navigation},
  author={Zhong, Linqing and Gao, Chen and Ding, Zihan and Liao, Yue and Ma, Huimin and Zhang, Shifeng and Zhou, Xu and Liu, Si},
  journal={arXiv preprint arXiv:2411.16425},
  year={2024}
}

@article{bai2025qwen2,
  title={Qwen2. 5-vl technical report},
  author={Bai, Shuai and Chen, Keqin and Liu, Xuejing and Wang, Jialin and Ge, Wenbin and Song, Sibo and Dang, Kai and Wang, Peng and Wang, Shijie and Tang, Jun and others},
  journal={arXiv preprint arXiv:2502.13923},
  year={2025}
}

@inproceedings{liu2024grounding,
  title={Grounding dino: Marrying dino with grounded pre-training for open-set object detection},
  author={Liu, Shilong and Zeng, Zhaoyang and Ren, Tianhe and Li, Feng and Zhang, Hao and Yang, Jie and Jiang, Qing and Li, Chunyuan and Yang, Jianwei and Su, Hang and others},
  booktitle={ECCV},
  year={2024},
}

@inproceedings{sam,
  title={Segment anything},
  author={Kirillov, Alexander and Mintun, Eric and Ravi, Nikhila and Mao, Hanzi and Rolland, Chloe and Gustafson, Laura and Xiao, Tete and Whitehead, Spencer and Berg, Alexander C and Lo, Wan-Yen and others},
  booktitle={CVPR},
  year={2023}
}

@inproceedings{lindenberger2023lightglue,
  title={Lightglue: Local feature matching at light speed},
  author={Lindenberger, Philipp and Sarlin, Paul-Edouard and Pollefeys, Marc},
  booktitle={CVPR},
  year={2023}
}

@inproceedings{he2017mask,
  title={Mask r-cnn},
  author={He, Kaiming and Gkioxari, Georgia and Doll{\'a}r, Piotr and Girshick, Ross},
  booktitle={CVPR},
  year={2017}
}

@article{chang2023goat,
  title={Goat: Go to any thing},
  author={Chang, Matthew and Gervet, Theophile and Khanna, Mukul and Yenamandra, Sriram and Shah, Dhruv and Min, So Yeon and Shah, Kavit and Paxton, Chris and Gupta, Saurabh and Batra, Dhruv and others},
  journal={arXiv preprint arXiv:2311.06430},
  year={2023}
}

@article{chaplot2020object,
  title={Object goal navigation using goal-oriented semantic exploration},
  author={Chaplot, Devendra Singh and Gandhi, Dhiraj Prakashchand and Gupta, Abhinav and Salakhutdinov, Russ R},
  journal={NeurIPS},
  year={2020}
}

@inproceedings{kwon2023renderable,
  title={Renderable neural radiance map for visual navigation},
  author={Kwon, Obin and Park, Jeongho and Oh, Songhwai},
  booktitle={CVPR},
  year={2023}
}

@inproceedings{ramakrishnan2022poni,
  title={Poni: Potential functions for objectgoal navigation with interaction-free learning},
  author={Ramakrishnan, Santhosh Kumar and Chaplot, Devendra Singh and Al-Halah, Ziad and Malik, Jitendra and Grauman, Kristen},
  booktitle={CVPR},
  year={2022}
}

@article{achiam2023gpt,
  title={Gpt-4 technical report},
  author={Achiam, Josh and Adler, Steven and Agarwal, Sandhini and Ahmad, Lama and Akkaya, Ilge and Aleman, Florencia Leoni and Almeida, Diogo and Altenschmidt, Janko and Altman, Sam and Anadkat, Shyamal and others},
  journal={arXiv preprint arXiv:2303.08774},
  year={2023}
}

@article{liu2023llava,
      title={Visual Instruction Tuning}, 
      author={Liu, Haotian and Li, Chunyuan and Wu, Qingyang and Lee, Yong Jae},
      journal={NeurIPS},
      year={2023},
}

@article{chen2023not,
  title={How To Not Train Your Dragon: Training-free Embodied Object Goal Navigation with Semantic Frontiers},
  author={Chen, Junting and Li, Guohao and Kumar, Suryansh and Ghanem, Bernard and Yu, Fisher},
  journal={RSS},
  year={2023},
}

@inproceedings{yokoyama2024vlfm,
  title={Vlfm: Vision-language frontier maps for zero-shot semantic navigation},
  author={Yokoyama, Naoki and Ha, Sehoon and Batra, Dhruv and Wang, Jiuguang and Bucher, Bernadette},
  booktitle={ICRA},
  year={2024},
}

@inproceedings{yu2023l3mvn,
  title={L3mvn: Leveraging large language models for visual target navigation},
  author={Yu, Bangguo and Kasaei, Hamidreza and Cao, Ming},
  booktitle={IROS},

  year={2023},

}

@inproceedings{zhang2024trihelper,
  title={Trihelper: Zero-shot object navigation with dynamic assistance},
  author={Zhang, Lingfeng and Zhang, Qiang and Wang, Hao and Xiao, Erjia and Jiang, Zixuan and Chen, Honglei and Xu, Renjing},
  booktitle={IROS},
  year={2024},
}

@article{long2024instructnav,
  title={Instructnav: Zero-shot system for generic instruction navigation in unexplored environment},
  author={Long, Yuxing and Cai, Wenzhe and Wang, Hongcheng and Zhan, Guanqi and Dong, Hao},
  journal={CoRL},
  year={2024}
}

@inproceedings{cai2024bridging,
  title={Bridging zero-shot object navigation and foundation models through pixel-guided navigation skill},
  author={Cai, Wenzhe and Huang, Siyuan and Cheng, Guangran and Long, Yuxing and Gao, Peng and Sun, Changyin and Dong, Hao},
  booktitle={ICRA},
  year={2024},
}

@inproceedings{krantz2023navigating,
  title={Navigating to objects specified by images},
  author={Krantz, Jacob and Gervet, Theophile and Yadav, Karmesh and Wang, Austin and Paxton, Chris and Mottaghi, Roozbeh and Batra, Dhruv and Malik, Jitendra and Lee, Stefan and Chaplot, Devendra Singh},
  booktitle={CVPR},
  year={2023}
}

@article{nie2025wmnav,
  title={WMNav: Integrating Vision-Language Models into World Models for Object Goal Navigation},
  author={Nie, Dujun and Guo, Xianda and Duan, Yiqun and Zhang, Ruijun and Chen, Long},
  journal={IROS},
  year={2025}
}

@inproceedings{sun2024prioritized,
  title={Prioritized semantic learning for zero-shot instance navigation},
  author={Sun, Xinyu and Liu, Lizhao and Zhi, Hongyan and Qiu, Ronghe and Liang, Junwei},
  booktitle={ECCV},
  year={2024},
}

@inproceedings{long2024discuss,
  title={Discuss before moving: Visual language navigation via multi-expert discussions},
  author={Long, Yuxing and Li, Xiaoqi and Cai, Wenzhe and Dong, Hao},
  booktitle={ICRA},
  year={2024},
}

@inproceedings{chen2024mapgpt,
  title={MapGPT: Map-Guided Prompting with Adaptive Path Planning for Vision-and-Language Navigation},
  author={Chen, Jiaqi and Lin, Bingqian and Xu, Ran and Chai, Zhenhua and Liang, Xiaodan and Wong, Kwan-Yee},
  booktitle={ACL},
  year={2024}
}

@inproceedings{liu2023bird,
  title={Bird's-eye-view scene graph for vision-language navigation},
  author={Liu, Rui and Wang, Xiaohan and Wang, Wenguan and Yang, Yi},
  booktitle={CVPR},
  year={2023}
}

@inproceedings{rajvanshi2024saynav,
  title={Saynav: Grounding large language models for dynamic planning to navigation in new environments},
  author={Rajvanshi, Abhinav and Sikka, Karan and Lin, Xiao and Lee, Bhoram and Chiu, Han-Pang and Velasquez, Alvaro},
  booktitle={Proceedings of the International Conference on Automated Planning and Scheduling},
  year={2024}
}

@inproceedings{szot2021habitat,
  title     =     {Habitat 2.0: Training Home Assistants to Rearrange their Habitat},
  author    =     {Andrew Szot and Alex Clegg and Eric Undersander and Erik Wijmans and Yili Zhao and John Turner and Noah Maestre and Mustafa Mukadam and Devendra Chaplot and Oleksandr Maksymets and Aaron Gokaslan and Vladimir Vondrus and Sameer Dharur and Franziska Meier and Wojciech Galuba and Angel Chang and Zsolt Kira and Vladlen Koltun and Jitendra Malik and Manolis Savva and Dhruv Batra},
  booktitle =     {NeurIPS},
  year      =     {2021}
}

@article{majumdar2022zson,
  title={Zson: Zero-shot object-goal navigation using multimodal goal embeddings},
  author={Majumdar, Arjun and Aggarwal, Gunjan and Devnani, Bhavika and Hoffman, Judy and Batra, Dhruv},
  journal={NeurIPS},
  year={2022}
}

@inproceedings{gadre2023cows,
  title={Cows on pasture: Baselines and benchmarks for language-driven zero-shot object navigation},
  author={Gadre, Samir Yitzhak and Wortsman, Mitchell and Ilharco, Gabriel and Schmidt, Ludwig and Song, Shuran},
  booktitle={CVPR},
  year={2023}
}

@article{krantz2022instance,
  title={Instance-specific image goal navigation: Training embodied agents to find object instances},
  author={Krantz, Jacob and Lee, Stefan and Malik, Jitendra and Batra, Dhruv and Chaplot, Devendra Singh},
  journal={arXiv preprint arXiv:2211.15876},
  year={2022}
}

@article{yadav2023ovrl,
  title={Ovrl-v2: A simple state-of-art baseline for imagenav and objectnav},
  author={Yadav, Karmesh and Majumdar, Arjun and Ramrakhya, Ram and Yokoyama, Naoki and Baevski, Alexei and Kira, Zsolt and Maksymets, Oleksandr and Batra, Dhruv},
  journal={arXiv preprint arXiv:2303.07798},
  year={2023}
}

@inproceedings{lei2024instance,
  title={Instance-aware exploration-verification-exploitation for instance imagegoal navigation},
  author={Lei, Xiaohan and Wang, Min and Zhou, Wengang and Li, Li and Li, Houqiang},
  booktitle={CVPR},
  year={2024}
}

@inproceedings{zhou2023esc,
  title={Esc: Exploration with soft commonsense constraints for zero-shot object navigation},
  author={Zhou, Kaiwen and Zheng, Kaizhi and Pryor, Connor and Shen, Yilin and Jin, Hongxia and Getoor, Lise and Wang, Xin Eric},
  booktitle={ICML},
  year={2023},
}

@article{team2024gemini,
  title={Gemini 1.5: Unlocking multimodal understanding across millions of tokens of context},
  author={Team, Gemini and Georgiev, Petko and Lei, Ving Ian and Burnell, Ryan and Bai, Libin and Gulati, Anmol and Tanzer, Garrett and Vincent, Damien and Pan, Zhufeng and Wang, Shibo and others},
  journal={arXiv preprint arXiv:2403.05530},
  year={2024}
}

@inproceedings{zhang2025multi,
  title={Multi-floor zero-shot object navigation policy},
  author={Zhang, Lingfeng and Wang, Hao and Xiao, Erjia and Zhang, Xinyao and Zhang, Qiang and Jiang, Zixuan and Xu, Renjing},
  booktitle={ICRA},
  year={2025},}

@article{sethian1999fast,
  title={Fast marching methods},
  author={Sethian, James A},
  journal={SIAM review},
  year={1999},
  publisher={SIAM}
}

@inproceedings{busch2025one,
  title={One map to find them all: Real-time open-vocabulary mapping for zero-shot multi-object navigation},
  author={Busch, Finn Lukas and Homberger, Timon and Ortega-Peimbert, Jes{\'u}s and Yang, Quantao and Andersson, Olov},
  booktitle={ICRA},
  year={2025},
}

@inproceedings{cao2025cognav,
  title={Cognav: Cognitive process modeling for object goal navigation with llms},
  author={Cao, Yihan and Zhang, Jiazhao and Yu, Zhinan and Liu, Shuzhen and Qin, Zheng and Zou, Qin and Du, Bo and Xu, Kai},
  booktitle=ICCV,
  year={2025}
}

@inproceedings{wu2024voronav,
  title={Voronav: Voronoi-based zero-shot object navigation with large language model},
  author={Wu, Pengying and Mu, Yao and Wu, Bingxian and Hou, Yi and Ma, Ji and Zhang, Shanghang and Liu, Chang},
  booktitle=ICML,
  year={2024}
}

@article{liu2025toponav,
  title={Toponav: Topological graphs as a key enabler for advanced object navigation},
  author={Liu, Peiran and Zhang, Qiang and Peng, Daojie and Zhang, Lingfeng and Qin, Yihao and Zhou, Hang and Ma, Jun and Xu, Renjing and Ji, Yiding},
  journal={arXiv preprint arXiv:2509.01364},
  year={2025}
}

@article{meta2024llama,
  title={Llama 3.2: Revolutionizing edge ai and vision with open, customizable models},
  author={Meta, AIa},
  journal={Meta AI Blog. Retrieved December},
  year={2024}
}

@article{liu2025drivepi,
  title={DrivePI: Spatial-aware 4D MLLM for Unified Autonomous Driving Understanding, Perception, Prediction and Planning},
  author={Liu, Zhe and Huang, Runhui and Yang, Rui and Yan, Siming and Wang, Zining and Hou, Lu and Lin, Di and Bai, Xiang and Zhao, Hengshuang},
  journal={CVPR},
  year={2026}
}

@inproceedings{li2026sgdrive,
  title={SGDrive: Scene-to-Goal Hierarchical World Cognition for Autonomous Driving},
  author={Li, Jingyu and Wu, Junjie and Hu, Dongnan and Huang, Xiangkai and Sun, Bin and Hao, Zhihui and Lang, Xianpeng and Zhu, Xiatian and Zhang, Li},
  booktitle=CVPR,
  year={2026}
}

@inproceedings{li2025imagidrive,
  title={ImagiDrive: A Unified Imagination-and-Planning Framework for Autonomous Driving},
  author={Li, Jingyu and Zhang, Bozhou and Jin, Xin and Deng, Jiankang and Zhu, Xiatian and Zhang, Li},
  booktitle=ICRA,
  year={2025}
}

@article{bi2025motus,
  title={Motus: A unified latent action world model},
  author={Bi, Hongzhe and Tan, Hengkai and Xie, Shenghao and Wang, Zeyuan and Huang, Shuhe and Liu, Haitian and Zhao, Ruowen and Feng, Yao and Xiang, Chendong and Rong, Yinze and others},
  journal={arXiv preprint arXiv:2512.13030},
  year={2025}
}

@article{lingbot-va2026,
  title={Causal World Modeling for Robot Control},
  author={Li, Lin and Zhang, Qihang and Luo, Yiming and Yang, Shuai and Wang, Ruilin and Han, Fei and Yu, Mingrui and Gao, Zelin and Xue, Nan and Zhu, Xing and Shen, Yujun and Xu, Yinghao},
  journal={arXiv preprint arXiv:2601.21998},
  year={2026}
}

@article{fast-wam,
  title={Fast-WAM: Do World Action Models Need Test-time Future Imagination?},
  author={Yuan, Tianyuan and Dong, Zibin and Liu, Yicheng and Zhao, Hang},
  journal={arXiv preprint arXiv:2603.16666},
  year={2026}
}
\newpage
\appendix
\section*{Appendix}
\setcounter{page}{1}

\section{Overview}
This supplementary material is organized as follows:

\begin{itemize}[leftmargin=*]
    \item Section~\textcolor{black}{\ref{sec:tasks}} provides the details of the three studied tasks.
    
    \item Section~\textcolor{black}{\ref{sec:exps}} provides details on the real-development experiment demo.
    
    \item Section~\textcolor{black}{\ref{sec:prompts}} details the prompts for VLM.
\end{itemize}

\section{Task introduce}\label{sec:tasks}
\begin{table}[h]
\centering
\renewcommand{\arraystretch}{1.2} 
\begin{tabular}{c|ccc}
\toprule 
\textbf{ON} & \textbf{Chair} & \textbf{Sofa} & \textbf{Plant} \\
\midrule 

\multirow{2}{*}{\textbf{IIN}} & 
    \includegraphics[width=0.2\textwidth]{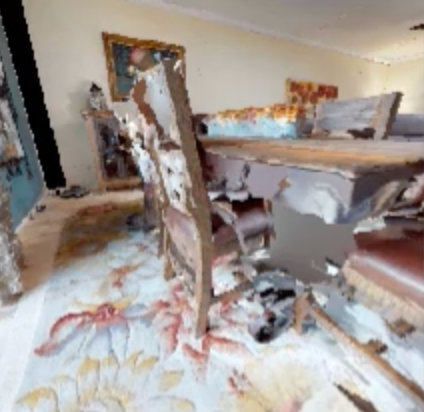} & 
    \includegraphics[width=0.2\textwidth]{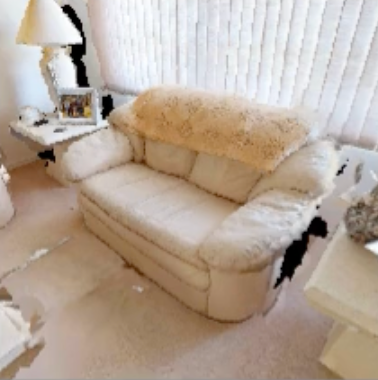} & 
    \includegraphics[width=0.2\textwidth]{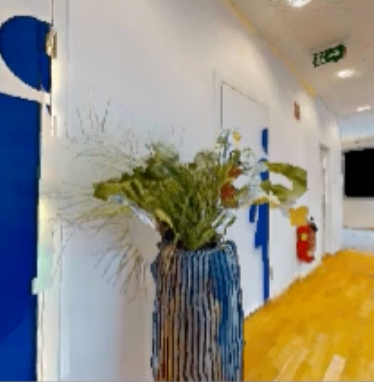} \\
    
& \textbf{Chair} & \textbf{Sofa} & \textbf{Plant} \\
\midrule

\textbf{TN} & 
    \begin{minipage}{0.2\textwidth}\small
The \textcolor{red}{\textbf{chair}} with a brown frame is positioned next to a table on a white carpet with red and yellow floral patterns in a room with beige walls.
\end{minipage} &
\begin{minipage}{0.2\textwidth}\small
The \textcolor{red}{\textbf{sofa}} is a plush, cream-colored loveseat with a soft, fluffy texture, topped with a cozy throw blanket in a bright, airy room.
\end{minipage} &
\begin{minipage}{0.2\textwidth}\small
The \textcolor{red}{\textbf{plant}} in a blue and white striped vase containing green foliage stands in a hallway with white walls and a yellow floor.
    \end{minipage} \\
\bottomrule 
\end{tabular}
\end{table}

We follow UniGoal~\cite{yin2025unigoal} provide samples of the goal in each task in Table 4 for better understanding. The goal for ON is a category in text format. The goal for IIN is an image with o located at the center. For TN, the goal is a description about o, such as its relationship with other relevant objects in the scene.

\section{Experiments}\label{sec:exps}
We provide additional simulation and real-world experiment results here. 
For simulation, we include video demonstrations on three tasks: IIN, TN, and ON. 
For real-world evaluation, our experiments are conducted on a Unitree Go2 quadruped robot, where the full \netName{} system runs on a Jetson Orin. 
Due to hardware limitations, we only use the onboard RGB camera of Go2 without relying on depth sensing for mapping. 
All experiments are conducted on a ThinkPad P16 laptop equipped with 32GB DDR5 memory and an NVIDIA RTX 3500 Ada Generation Laptop GPU with 12GB VRAM. Action commands are generated by Qwen2.5-VL-7B.
\textbf{We provide video demonstrations for TN tasks, which are included in the supplementary material as a compressed file.}

\section{Prompts details}\label{sec:prompts}
For clarity, we provide detailed prompt templates used in our system.

\begin{table*}[t]
\centering
\caption{Prompts used in our Agent.}
\begin{tabularx}{\textwidth}{lX}
\toprule
\textbf{Prompt Name} & \textbf{Description} \\
\midrule

Goal Room Recognition &
Analyze the image and estimate the most likely room types.
The output must be a JSON list with keys \texttt{room} and \texttt{probability}.
The room names must be selected from a predefined list.

Example output:
\texttt{[ \{"room": "bedroom", "probability": 0.95\} ]}

\\

Observation Room Recognition &
Given an image, output the single most likely room type.
The answer must be selected from a predefined room list and contain
only the room name without additional text.

Example output: \texttt{kitchen}

\\

Instance Verification &
Given a goal image and the current observation,
determine whether the same object instance appears in both images.
The output must be a JSON object with key \texttt{is\_match}.

Example output:
\texttt{\{"is\_match": true\}}

\\
\bottomrule
\end{tabularx}
\end{table*}

\begin{table*}[t]
\centering
\caption{Prompt templates used in cognitive map.}
\begin{tabularx}{\textwidth}{lX}
\toprule
\textbf{Prompt} & \textbf{Description} \\
\midrule

Image-to-Object Detection &
Given an indoor scene image and a candidate object list, the model returns a JSON object containing the visible objects selected only from the candidate list. The target object must always appear in the output.
Example: \texttt{\{"objects": ["chair","sofa","table"]\}} \\

Key Landmark Object Generation &
The model analyzes the image and selects the top 3--5 most important landmark objects useful for robot navigation, filtering out generic background items (e.g., wall, floor). Output format: \texttt{\{"key\_objects": ["sofa","armchair","lamp"]\}} \\

Target Attribute Extraction &
Given an image and a target object name, the model produces intrinsic attributes (appearance, shape, color, material) and extrinsic attributes (surrounding objects and environment). Output is a JSON object describing these attributes. \\

Candidate Similarity Estimation &
Given a textual goal description and candidate object images, the model evaluates how well the candidate matches the target. It outputs intrinsic and extrinsic similarity scores in the range [0,1] with a brief reasoning. \\

Target Presence Inference &
Based on contextual objects observed in the scene, the model infers the likelihood that the target object exists nearby but is currently unseen. The output includes a likelihood score and a short explanation. \\

\bottomrule
\end{tabularx}
\end{table*}

\end{document}